%% file: ICRA2022.tex
\documentclass[letterpaper, 10 pt, conference]{ieeeconf}  

\IEEEoverridecommandlockouts                              

\usepackage[utf8]{inputenc} 

\usepackage[bookmarks=true]{hyperref}

\usepackage{mathtools}

\usepackage{graphicx}
\usepackage{amsmath,amssymb,stmaryrd,mathtools}

\usepackage{url}            
\usepackage{booktabs}       
\usepackage{amsfonts}       
\usepackage{nicefrac}       
\usepackage{microtype}      
\usepackage{balance}
\usepackage{graphicx}
\usepackage{placeins}
\usepackage{booktabs}
\usepackage{multirow}
\usepackage{float}
\usepackage{xspace}
\usepackage{svg}
\hypersetup{
    colorlinks=true,
    linkcolor=black,
    filecolor=magenta,      
    urlcolor=cyan
}
\usepackage{capt-of,etoolbox}

\input{commands}

\input{math_commands}

\usepackage{amsmath}
\usepackage{mathtools}
\usepackage{wrapfig}

\title{Visual Affordance Prediction for Guiding Robot Exploration} 

\author{Homanga Bharadhwaj, Abhinav Gupta, Shubham Tulsiani 
\thanks{The authors are with the Robotics Institute, Carnegie Mellon University. Link to project website \url{https://sites.google.com/andrew.cmu.edu/affordance-robotics/home} }}

\begin{document}

\maketitle
\thispagestyle{empty}
\pagestyle{empty}

\maketitle

\begin{abstract}
Motivated by the intuitive understanding humans have about the space of possible interactions, and the ease with which they can generalize this understanding to previously unseen scenes, we develop an approach for learning visual affordances for guiding robot exploration. Given an input image of a scene, we infer a distribution over plausible future states that can be achieved via interactions with it. We use a Transformer-based model to learn a conditional distribution in the latent embedding space of a VQ-VAE and show that these models can be trained using large-scale and diverse passive data, and that the learned models exhibit compositional generalization to diverse objects beyond the training distribution. We show how the trained affordance model can be used for guiding exploration by acting as a goal-sampling distribution, during visual goal-conditioned policy learning in robotic manipulation.
\end{abstract}

\section{Introduction}
Consider the images $o_g$ in Figure 1. We humans can effortlessly understand the depicted scenes e.g. a bottle lying on the table in the top image, or a teddy lying next to a pot. More importantly, we can infer not only what \emph{is}, but also what \emph{can be} e.g. one can imagine the bottle being placed at a different location on the table, or turned to lie horizontally, or perhaps taking the teddy to place it inside the pan. We instinctively make such judgements across a myriad of everyday scenarios, understanding that an open cabinet can be closed, or that an egg can be broken, or spilled liquid wiped away. In this work, our goal is to build a computational system with similar capabilities, which can be used directly as a goal-sampling distribution for robot exploration. Given just a single image depicting generic (possibly unseen) objects, we wish to predict possible configurations that may occur as a result of a human (or a robot) interacting in the scene.

We are inspired by Gibson’s seminal work on \emph{affordances} which argues for developing intelligent agents with an intuitive understanding of possible interactions they can have with their environment~\cite{gibson1979ecological}. Initial attempts in the vision community formalized this as a pixelwise labeling task (e.g. `sittable’ surface), but these do not explicitly model the actions or their effects~\cite{fouhey2012semantic,wang2017binge}. An alternative approach has been to use geometric models~\cite{Gupta11,fouhey2012people} to predict human-centric affordances. Towards a richer parametrization, recent approaches have pursued `visuo-motor’ affordances where the space of possible interactions is modeled via predicting possible low-level actions that an agent can execute to affect its environment~\cite{nair2018visual,pong2019skew,mo2021where2act,corl2020roll}. While such visuo-motor affordances can be directly translated to agent behavior, we argue that the requirement of inferring precise actions associated can be restrictive e.g. babies may understand that a fruit can be cut even if they can’t do it themselves, as perhaps an adult may know that a bike’s tire can be removed even if not sure how to precisely do so. In addition, training models for such visuo-motor affordances requires annotated data, which can be restrictive to scale for diverse settings. In our work, we therefore pursue `visual affordances’, where instead of modeling the action trajectories, we aim to model what their \emph{effects} can be. More concretely, given a single input image, we pursue the task of conditionally generating diverse and plausible images of the same scene that can be achieved as a result of an agent’s interaction.

This task of inferring visual affordances is an increasingly common one in the robot learning community, where an understanding of interesting plausible actions can help guide exploration. However, the typical methods only learn domain-specific affordances — relying on active interaction in a specific (lab) setting, they learn models for interactions with the particular objects in the particular environment~\cite{nair2018visual,pong2019skew,khazatsky2021can,nair2020contextual}. In addition to their limited generalizability, these active interaction-based methods also do not learn more complex behaviors (e.g. stacking blocks) as random interactions from an agent are unlikely to lead to such interesting transitions. Our key insight in this work is that instead of only relying on active interactions with limited variation, such visual affordances can be learned from large-scale diverse  passive data. Just as we humans can learn from watching others (e.g. instructional videos), our system learns visual affordances using interaction videos depicting generic human and robot interactions across varied settings. Using this trained affordance model, when a robot is placed in a certain scene, we can generate goals corresponding to plausible manipulations of the objects in the scene, and learn a goal-conditioned policy through self-supervised exploration.


Specifically, we consider the setting of exploration in goal-conditioned policy learning and use the affordance model for sampling goals that the policy tries to reach during training. The affordance model consists of two stages: 1) we learn discrete latent embeddings from images by training a VQ-VAE~\cite{vqvae} on the entire dataset 2) we learn conditional prediction of latent codes through an auto-regressive generation procedure using a Transformer architecture~\cite{chen2020generative}. Because of the diversity seen in training, the affordance model is able to generalize to new environments and unseen objects, and predict interesting plausible configurations in settings never seen in training. Due to the diversity of sampled goals, the policy is incentivized to explore broadly and learn interesting behaviors like stacking, compared to curiosity-based exploration strategies.

\begin{figure*}[t]
    \centering
    \includegraphics[width=\textwidth]{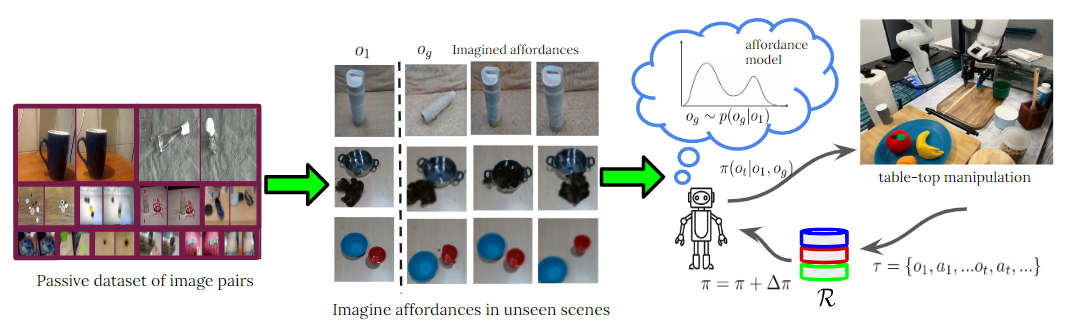}
    \caption{\small{We develop an approach for goal-directed robot exploration by learning a goal-sampling distribution followed by self behavior cloning with exploration trajectories. The goal-sampling distribution is a visual affordance model trained from large passive datasets of image pairs to predict a distribution over goal images given an initial image of the scene. We show that this affordance model enables goal-directed robot exploration, for learning diverse behaviors like pushing, grasping, and stacking in a table-top manipulation setting.}}
    \label{fig:teaser}
\end{figure*}

\section{Related Works}

\noindent\textbf{Actionable Information with Visual Observations.}
Prior work has explored the problem of learning how to interact with objects in the scene from visual observations, in different settings like dextrous manipulation~\cite{graumandextrous} and mobile navigation~\cite{graumaninteraction}. These approaches typically use either passive observations (for example human videos)~\cite{graumanvideo,brahmbhatt2019contactdb,fouhey2012people} or active interactions with a robotic agent~\cite{pinto2016supersizing}. Some prior works also leverage robot simulators to randomize the generation of different types of objects in order to learn affordances for tasks like grasping~\cite{khansari2020action} and pushing~\cite{zeng2018learning} objects. The unifying idea in all these approaches is learning \textit{how} to interact with objects in the scene. They tackle a problem slightly orthogonal to ours, where by learning affordances, we refer to visualizing the \textit{result} of interactions, and not specifically how to obtain those interactions. \\

\noindent\textbf{Generative Modeling.} There have been significant recent advances in deep generative models, with high-quality realistic images being produced by StyleGAN~\cite{stylegan1,stylegan2} and BigGAN~\cite{biggan,bigbigan} among others that use transformers~\cite{vqvae,tulsiani2021pixeltransformer,imagetransformer,chen2020generative,esser2021taming}. While most of these models tackle the problem of unconditional generation, a closely related problem to our setting is that of conditional image generation. This is typically defined as the task of generating new images from a dataset conditional on certain class attributes~\cite{sohn2015learning,limitationsconditional}. Video generation methods~\cite{ccvs} using transformers perform per-timestep prediction, whereas our affordance model samples (long-horizon) goal images that we show are useful for robot exploration. Some style transfer approaches~\cite{pix2pix,cyclegan} also have a similar conditional generation formulation, where the conditioning is done on a source image as opposed to a class attribute. Our affordance model is an approach for better conditioning on images, where we want the generations to capture configurational changes in the attributes of objects in the source image. \\

\noindent\textbf{Robot Exploration.} Generative models have been used for goal-sampling in visual robot learning~\cite{nair2018visual,nair2020contextual,pong2019skew,lin2019reinforcement}. Learning a good goal-sampling strategy for diverse goals and training goal-conditioned policies to try and reach sampled goals is a paradigm for exploration without the use of explicit heuristics like curiosity~\cite{pathak2017curiosity,episodic} or reduction of uncertainty~\cite{rnd,osband1}. Most prior work in this vein has learned goal-sampling models only in the experiment domain either by collecting large robot interaction datasets, or training the goal-sampler online jointly with policy learning~\cite{nair2018visual,lin2019reinforcement,pong2019skew,corl2020roll}. Collecting large datasets in the lab is expensive, and the goal-sampling models trained solely on a specific-lab setting is unlikely to generalize to other settings without requiring re-training. In contrast to these approaches we learn a model for generating visual affordances from various internet data - \textit{not} data collected in the experimenter's lab. We show how the learned affordance model can be used to guide exploration for goal-conditioned policy learning in environments with unseen objects.

 \begin{figure*}[t]
    \centering
    \includegraphics[width=\textwidth]{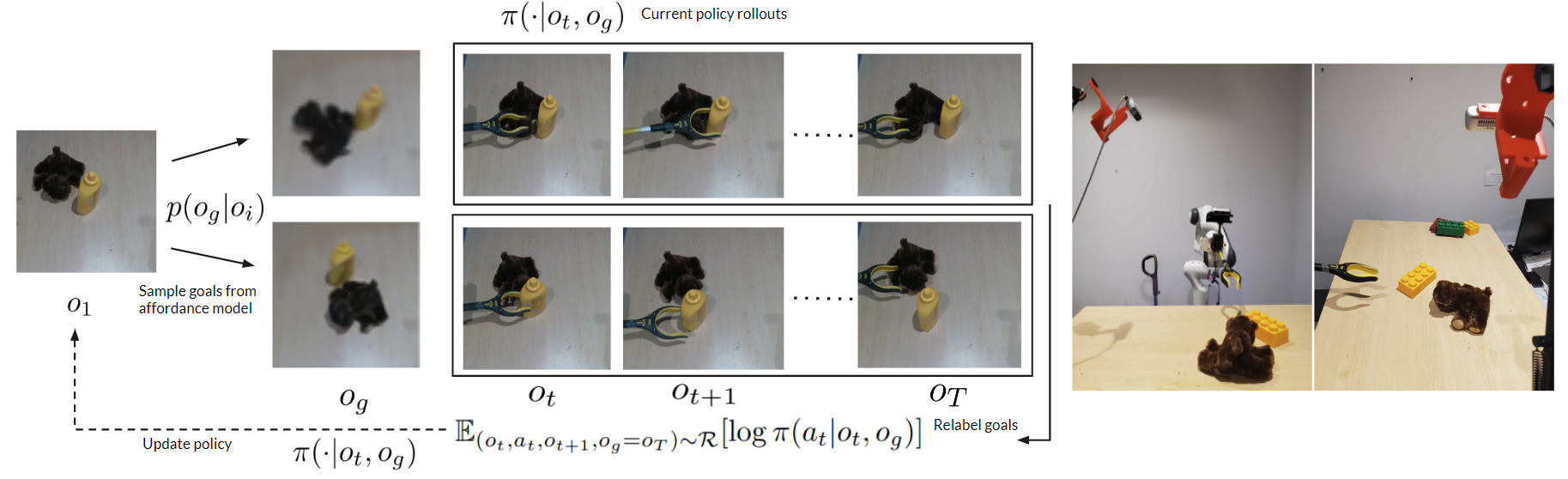}
    \caption{\footnotesize Affordance-driven exploration and policy learning through hindsight goal re-labeling. Given an initial configuration, we sample a goal with the affordance model, execute rollouts with the current policy, and store the transitions in the replay buffer. For updating the policy, we sample transitions from the replay buffer and re-label goals to be the final frames in the corresponding trajectory. The process is described in detail in section~\ref{sec:robotlearning} On the right we show two different views of the robot workspace, with a Franka Panda arm and an overhead Realsense camera.}
    \label{fig:bc}
\end{figure*}

\section{Approach}
We develop an approach for learning visual affordances from passive data that can be used as goals for guiding robot exploration. We use the term affordance to mean the set of possible interactions in a particular scene. Given an  image of a scene, we want the affordance model to generate a new image with different configurations of the same objects in the scene. This model can be used for goal-directed exploration, such that in a new scene, the agent can sample a goal conditioned on the current image, and execute actions to try and reach the sampled goal.



Since collecting data in-lab is expensive due to a lot of manual effort, we make use of the rich diversity of passive data existing in the internet, to learn the affordance model, and use it directly for goal-sampling. The affordance model is two-staged: we first learn a VQ-VAE~\cite{vqvae} to discretize the space of continuous images, and then learn a transformer based auto-regressive model that (in latent space) predicts a possible goal given the input. We then project the generated latent goal, back to the image-space with the decoder of the VQ-VAE. This framework lets us both: a) generate diverse goals, and b) produce realistic and high-resolution images. %


For goal-conditioned policy learning, self-supervised approaches typically adopt a learning pipeline consisting of two (often inter-leaved) components: 1) exploration to collect data of interactions with the environment, and 2) policy learning using the interaction trajectories. We show that it is possible to drive better exploration by using the trained affordance model as a conditional goal-sampling distribution that samples diverse realistic goals in an unseen scene.

Concretely, we learn a goal-conditioned policy through exploration, where the goals are sampled from a learned affordance model in the beginning of the trajectory. In order to learn such a policy through self-supervised exploration, the sampled goals must be \textit{plausible} and \textit{diverse}, corresponding to different arrangement of the objects in the scene. Depending on the objects in the scene, the different arrangements could be putting a lid on a pot, pushing a cup across the table, grasping an object and rotating it etc. We show that such behaviors emerge from the trained affordance model, and describe how we can learn a goal-conditioned policy through hindsight re-labeling of exploration trajectories.

In the next sub-sections, we first describe the architecture and training details of the affordance model, and then discuss how we use the model as a goal-sampler for guiding robot exploration in manipulation tasks.

\subsection{Learning to Imagine Goals}

Instead of modeling in the image-space directly, in order to reduce spatial redundancies and implicitly abstract out interesting objects in the scene, we first learn a down-sampled encoding of high resolution images. To learn these discrete latent embeddings, we employ a VQ-VAE to learn lossy encoder and decoder models that can transform latents to generate realistic image samples. Having a discrete latent space allows flexibility in modeling features (such as objects in the scene) that span multiple pixels in the image, and do not encode local imperceptible differences that are not significant. 
\\

\noindent\textbf{Training details of the VQ-VAE.} Let the images be $o\in\mathbb{R}^{H\times W\times 3}$. The VQ-VAE model consists of an encoder $E(o) \in \mathbb{R}^{h\times w\times L}$, a codebook $C=\{e_k\}_{k=1}^K$ with elements $e_i$ of size $L$ and a decoder $D(e)\in\mathbb{R}^{H\times W\times 3}$. The quantization happens in the channel space, where all the $h\times w$ vectors of size $L$ are replaced by their nearest codebook vectors $e_i$, and the resulting quantized latent code of dimension $h\times w \times L$ is fed to the decoder. The training objective for this model, following prior work~\cite{vqvae} is as follows:
\begin{align*}
    \gL_{\text{vqvae}} = \E_{o\sim\gD}\left[ || o -  D(e)||_2 + ||E(o) - \text{sg}[e]||_2 \right]
\end{align*}
Here, sg refers to the stopgradient operator, that is defined as identity during forward-computation.

After training the VQ-VAE, corresponding to each image $o_c$, we obtain a downsampled encoding $z_c$. We then learn an auto-regressive prior using a Transformer architecture to model $p(z_g|z_c)$. Finally, with the decoder of the VQ-VAE we obtain a corresponding image $o_g$ from the latent $z_g$. Learning this conditional generation model in the latent space with a powerful Transformer architecture enables us to achieve compositional generalization, by generation of plausible goals in scenes where there are multiple objects. The stochasticity of the model is helpful in ensuring that the generated samples are allowed to be diverse.\\

\begin{figure}[t]
    \centering
    \includegraphics[width=\linewidth]{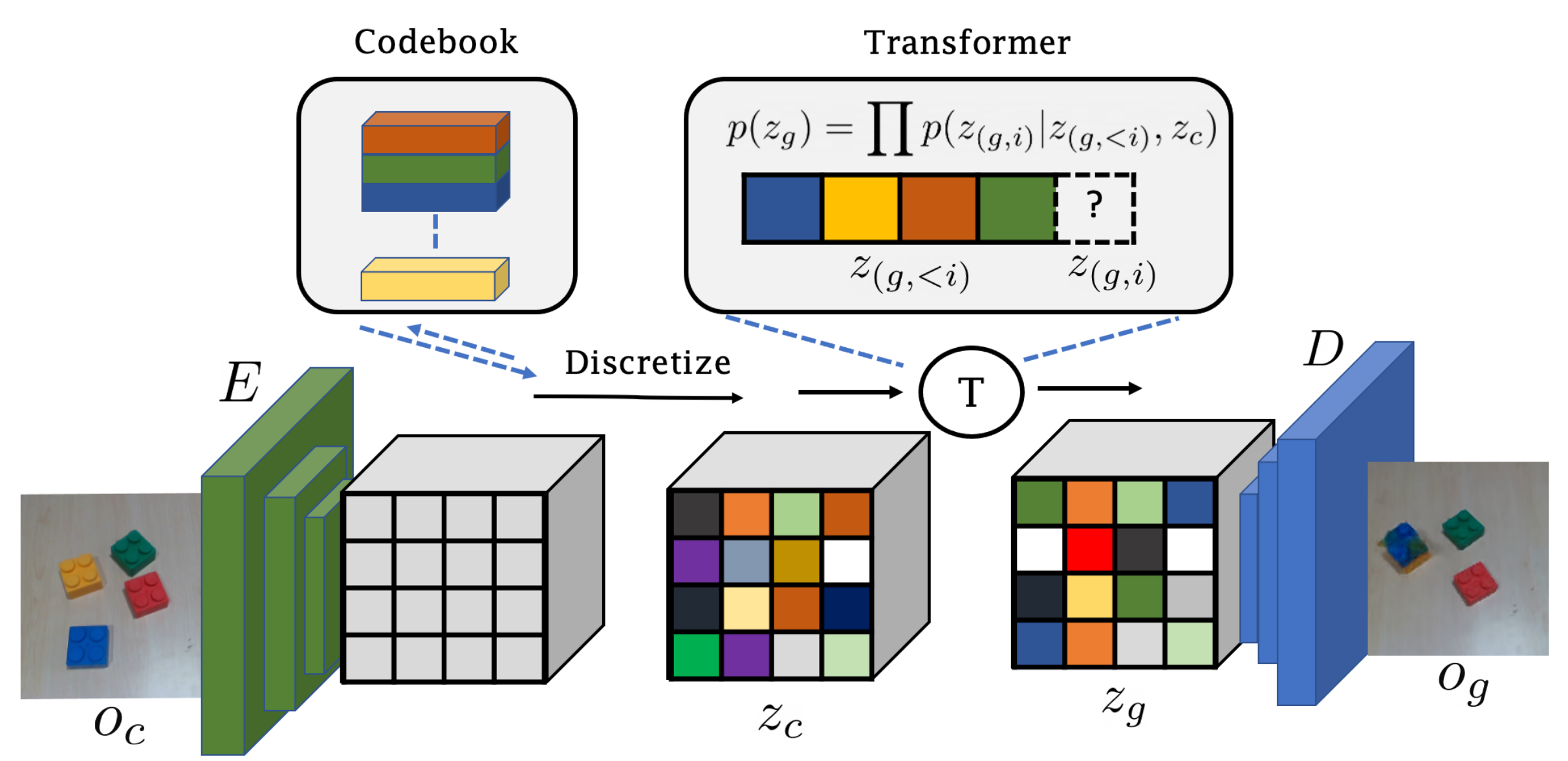}
    \caption{\footnotesize Illustration of a forward pass through the affordance model. The initial scene $o_c$ is first encoded by a VQ-VAE encoder, and converted to a discretized latent code by swapping nearest neighbor encodings from a codebook $C$. The resulting discrete latent code $z_c$ is passed through the Transformer. The autoregressive generation yields a latent code $z_g$ which is decoded by the VQ-VAE decoder to a plausible goal image $o_g$. }
    \label{fig:main}
\end{figure}

\noindent\textbf{Auto-regressive goal sampling.} In Fig.~\ref{fig:main}, we show a forward pass through our affordance model that generates a goal image $o_g$, given a conditioning image $o_c$. The initial image $o_c$ is first encoded by the VQ-VAE to a discrete latent code $z_c$. The Transformer then generates the latent code corresponding to the goal image, $z_g\sim p(z_g|z_c)$. The generation process happens pixel by pixel in raster-scan order and channel by channel for one pixel.  Hence we can denote the auto-regressive generation process as:
\begin{align*}
    p(z_g) = \prod_i p\left(z_{(g,i)}|z_{(g,<i)},z_c\right)
\end{align*}
This equation shows that the prediction of the $i^{\text{th}}$ pixel in the latent code $z_{(g,i)}$ is conditioned on the already predicted pixels $z_{(g,<i)}$ and the initial latent code $z_c$. \\

\noindent\textbf{Learning objective.} We instantiate the affordance model as $p_\psi(o_g|o_c)$ and aim to maximize $\E_{(o_g,o_c)\sim\gD}[\log p_\psi(o_g|o_c)]$. Given image pairs $(o_c,o_g)\sim\gD$, we transform each image to their respective quantized latent code $z_c$ and $z_g$. The autoregressive transformer model is used to generate $\hat{z}_g$ conditioned on $z_c$, i.e. $\hat{z}_g\sim p(z_g|z_c)$. We train the model to maximize  $\E_{(z_g,z_c)\sim\gD}[\log p(z_g|z_c)]$. \\

\noindent\textbf{Training Details of the Transformer.} For the above generation, we use an architecture similar to Image-GPT~\cite{chen2020generative}. It consists of an encoder-decoder model with masked convolutions and self-attention layers in the decoder. The auto-regressive model is preserved by padding the input, and replacing the paddded pixel values with the generated values and repeating the process recursively. Finally, all the values in $z_g$ correspond to valid generations that become more and more accurate as training progresses, and so we eventually have a model of the form $p(z_g|z_c)$. We then feed in $z_g$ to the VQ-VAE decoder, obtaining the decoded image frame $o_g$. This completes the structure of the affordance model $p_\psi(o_g|o_c)$.  After training the model, given a new initial image $\hat{o}_c$, we can sample a plausible goal $o_g\sim p_\psi(o_g|\hat{o}_c)$. We empirically analyze this model in the next section, and show results of goal generations in unseen scenes.

\subsection{Affordance-driven Robot Exploration}
\label{sec:robotlearning}

We use the affordance model trained solely on passive data for goal-conditioning and do not fine-tune on any lab-specific data. Because we trained the model on diverse passive data, we observe good generalization for affordance prediction in the robot learning setting. We obtain exploration trajectories with interesting behaviors, and are eventually able to learn better goal-conditioned policies compared to alternate exploration strategies like curiosity~\cite{pathak2017curiosity} that learn from scratch in the environment, or those that sample goals from other sampling distributions like a VAE-prior~\cite{nair2018visual}.

Given the affordance model, we train a policy that chooses actions for executing a particular task specified by a goal image. We refer to the observation the robot sees at time-step $t$ of an episode as $o_t$, and denote the goal image as $o_g$.  We denote the goal-conditioned policy as $\pi(a_t|o_t,o_g)$. We learn this policy by simple behavior cloning with goal-relabeling. After sampling a goal $o_g\sim p_\psi(o_g|o_1)$ and executing a trajectory $(o_1,a_1,o_2,a_2,...,o_T;o_g)$ to reach the goal, even if the final configuration $o_T$ is not similar to the goal configuration, the executed trajectory provides useful information for reaching the configuration the system ended up in. We describe the overall framework in Fig.~\ref{fig:bc}. During deployment, test goal images are sampled from some distribution $o_g\sim p(\gG)$, and the robot must interact with the objects in the scene to reach the goal configuration.  \\

\noindent\textbf{Training the goal-conditioned policy.} Given observation $o_1$ corresponding to the initial scene, we sample a goal from the affordance model $o_g\sim p_\psi(o_g|o_i)$ and execute a trajectory $(o_1,a_1,o_2,a_2,...,o_T;o_g)$, which we store in the replay buffer $\gR$. In Fig.~\ref{fig:bc} we see examples of goals sampled in a scene. The affordance model enables sampling diverse goals and encourages the policy to try and reach them, thereby ensuring interesting interactions. Given the data of interactions in the replay buffer $\gR$, we perform goal re-labeling and use tuples $(o_t,a_t,o_{t+1},o_g=o_T)$ to update the policy $\pi(a_t|o_t,o_g)$. This corresponds to a variation of hindsight-experience replay, that has been shown to be useful in prior works~\cite{andrychowicz2017hindsight,ghosh2019learning,li2020generalized}. The rationale for this is that during exploration, although the incompletely trained policy might not reach the sampled goal, the final state it reaches is still a \textit{potential} goal, and the executed sequence of actions provides guidance on reaching this goal. The policy update method is to do simple behavior cloning that maximizes the the likelihood $\E_{(o_t,a_t,o_{t+1},o_g=o_T)\sim\gR}[\log\pi(a_t|o_t,o_g)]$. Note that we do not need samples to be on-policy for this update and so we can interleave exploration of a few trajectories with a policy update phase where we randomly sample tuples from different trajectories in the replay buffer.

\section{Experiments}
\label{sec:experiments}

Through experiments, we aim to understand the following research questions:

\begin{itemize}
\item How effective is the affordance model in generating diverse scenes with plausible object manipulations?
\item How effective are the generated plausible and diverse affordances in guiding robot exploration?
\end{itemize}

\subsection{Setting}
\label{sec:setting}

\noindent\textbf{Data.} For training the affordance model, we use data from three different internet sources: somethingsomethingv2~\cite{smth}, Berkeley~\cite{khazatsky2021can}, and JHU CoStar~\cite{jhu}. We did not collect any additional data ourselves, and use only pairwise frames extracted from these datasets for training. We use the trained visual affordance model as-is for robot experiments and do not do any fine-tuning in the lab.  

\noindent\textbf{Baselines.} We compare our model against two relevant baselines, a Conditional VAE (CVAE)~\cite{sohn2015learning}, and a Conditional GAN (Pix2Pix)~\cite{pix2pix} for conditional generation and robot exploration, and in addition with a curiosity baseline~\cite{pathak2017curiosity} for robot exploration. We train these models on the same data as our model, and perform qualitative evaluations as well as quantitative comparisons through several metrics described in the next sub-section. After analyzing the visual generations of the baselines in section~\ref{sec:metrics}, we evaluate policy learning behaviors in section~\ref{sec:policy}
 \begin{figure}[h!]
    \centering
    \includegraphics[width=\columnwidth]{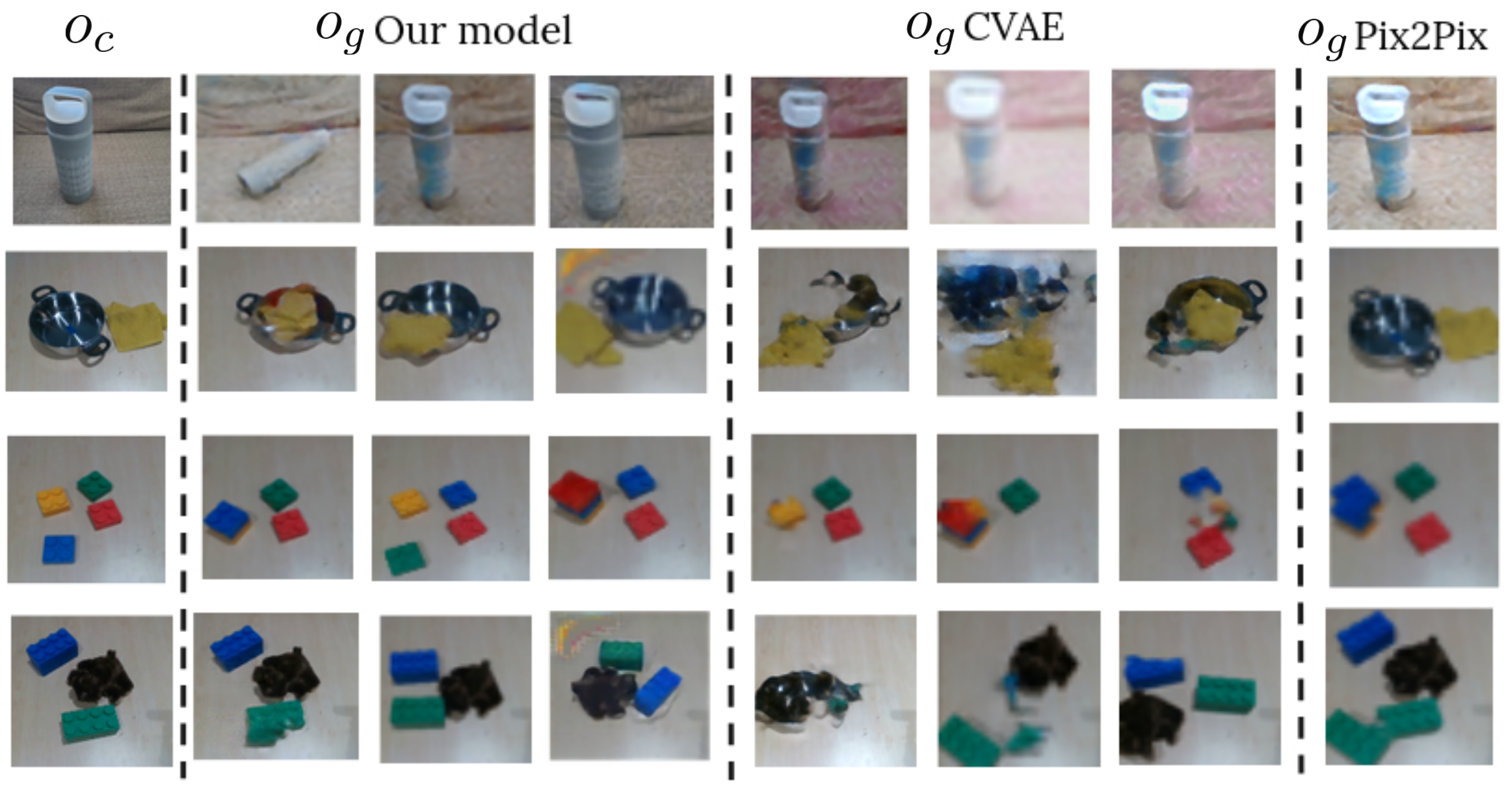}
    \caption{\footnotesize Qualitative analysis of the proposed affordance model with the baselines CVAE and Pix2Pix. Given initial conditioning frame $o_c$ that corresponds to an unseen configuration during training, we see that the sampled affordances $o_g$ from our model are diverse and correspond to plausible interactions in the scene. For the baselines, we see that the generated frames are not diverse and sometimes omit certain objects from the scene or introduce different artifacts.}
    \label{fig:affordances}
\end{figure}

\subsection{Evaluating Predicted Affordances}
\label{sec:metrics}
As highlighted by prior work, evaluating the quality of synthesized images is challenging~\cite{pix2pix,cyclegan}. Typical metrics used in assessing image reconstruction (like the pixel mean-squared error) do not translate well to assessing the quality of novel image generations. Further, these metrics will not be useful for understanding how diverse and plausible are the generated affordances for novel scenes. Hence we consider a metric based on \textit{human perceptual evaluation}. Motivated by evaluation protocols in prior works~\cite{gupta2017characterizing,pix2pix,cyclegan,woods2015conducting}, we conduct an perceptual study with Amazon Mechanical Turk (MTurk). We run the MTurk perceptual study, following the protocol from prior work~\cite{pix2pix,cyclegan} where images are shown on the screen for 3 seconds and workers are asked to guess certain proprieties of the images. In our setting, we show three images per screen and ask the workers to choose one among the two rightmost images. We mention in the instructions that the task is to guess which of the two images on the right correspond to plausible manipulations of objects on the left image. We randomize the ordering of images such that one image is from our model and the other is from a baseline and compare the average number of times workers choose ours compared to the baselines.


\noindent\textbf{Qualitative results.} In Fig.~\ref{fig:affordances} we perform a qualitative analysis of the proposed affordance model with the baselines CVAE and Pix2Pix.  For each of the three initial scenes $o_c$ in the first column, we generate three affordances with our model, three with the CVAE model, and one with the Pix2Pix model (since the model is not stochastic). The initial scenes are unseen during training. We can see that the affordances generated by our model are diverse and involve plausible manipulations of the different objects in the scene. We observe the emergence of interesting affordances like stacking of blocks and re-orientation of the bottle. Whereas for the baselines, sometimes certain objects are omitted from the scene, are non-realistically generated, and do not involve diverse manipulations of the objects. Given that the objects have never appeared in the training data before, this provides evidence that our method can generalize to new scenes and compose affordances for different objects.

\noindent\textbf{Quantitative results.}
In Table~\ref{tb:turkresults}, we report results from the MTurk perceptual study. 40 workers participated in our study. In each trial, workers saw an image on the left, and were asked to choose which of the two images on the right corresponded to plausible manipulations of objects on the left image. The numbers show the $\%$ of trials where workers chose a sample from our model as opposed to that from the baselines, averaged over all the workers. Since the numbers are $>50\%$, we see that workers preferred samples from our model compared to the baselines. This confirms the plausibility of our generated affordances.


\setlength{\tabcolsep}{8pt}
\begin{table}[h!]
\centering
\caption{\footnotesize Pairwise comparison of results for the MTurk perceptual study. We tabulate the $\%$ of trials where workers chose a sample from our model as opposed to that from the baselines, averaged over 40 workers who participated in our study. Higher is better. A number $>50\%$ shows that workers chose our model's sample more number of times compared to the baselines.}
\begin{tabular}{@{}ccccc@{}}
\toprule
              & \textbf{Pix2Pix} & \textbf{CVAE} \\ \midrule
\textbf{Our}  &    $69.8\pm 11.9$    &    $75.5\pm 10.8$               \\ \bottomrule
\end{tabular}
\label{tb:turkresults}
\end{table}


 

\subsection{Benchmarking Affordance-Guided Exploration}

\begin{figure}[h!]
    \centering
\includegraphics[width=\linewidth]{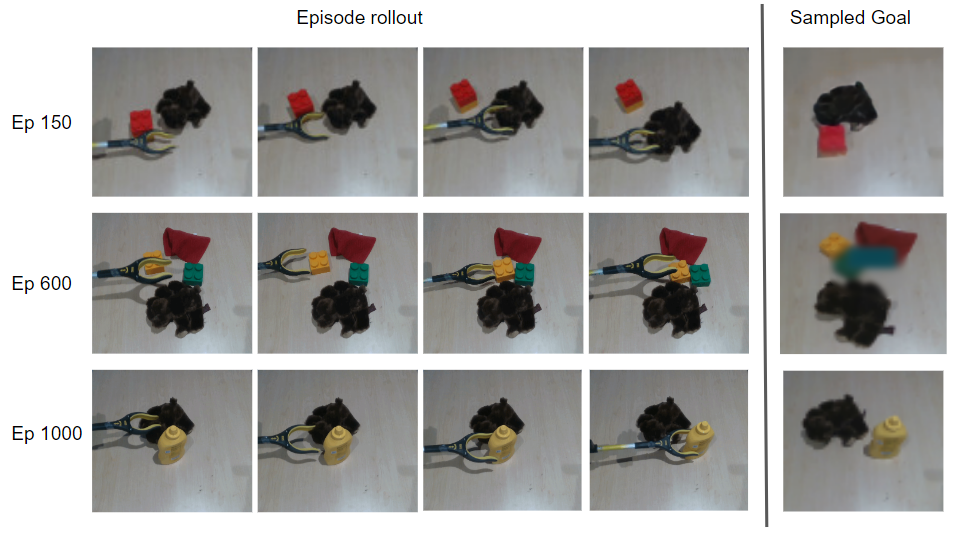}
    \caption{\footnotesize We show visualizations of exploration during training, corresponding to different episodes. On the right, we show the respective goal sampled from the affordance model. Towards the end, we see the evolution of interesting behaviors like grasping and the policy leading to behaviors that reach the goal image.
}
    \label{fig:explorationviz}
\end{figure}

\label{sec:policy}

\setlength{\tabcolsep}{8pt}
\begin{table}
\caption{\footnotesize We show results of tests for various robot manipulation tasks. For each task, we have two test goal configurations, and report success rate over 10 trials each.  }
\label{tb:robot}
\centering
\begin{tabular}{cccc}
\hline
                  & \textbf{Pushing} & \textbf{Pick and Place} & \textbf{Stacking} \\ \hline
\textbf{Curiosity} & 50\%             & 40\%                    & 30\%              \\
\textbf{CVAE} & 30\%             & 20\%                    & 10\%              \\
\textbf{Pix2Pix} & 30\%             & 10\%                    & 10\%              \\
\textbf{Our}      & 70\%             & 60\%                    & 60\%              \\ \hline
\end{tabular}

\end{table}

\begin{figure}[h!]
    \centering
    \includegraphics[width=1.03\linewidth]{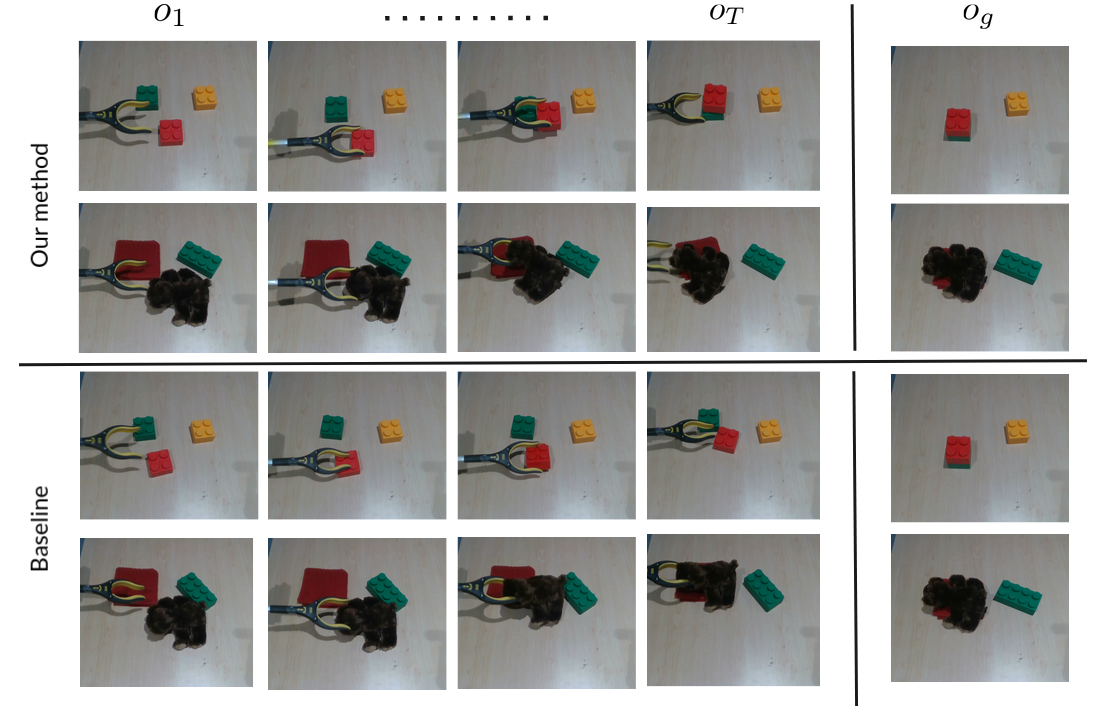}
    \caption{\footnotesize We show examples of two evaluation runs of our method and the curiosity baseline corresponding to two test goals in stacking and pick and place respectively. On the right are the goal images that the policy is conditioned on for evaluation, and the sequence of observations $o_1,...,o_T$ corresponds to the executed trajectory.}
    \label{fig:robotexecutions}
\end{figure}

 In this section, we empirically analyze the framework in terms of generating diverse affordances that serve as goals for aiding exploration in robot policy learning. We consider a goal-directed policy learning setting, where the robot needs to set its own goals, and explore the environment to try and reach those goals. There is no notion of \textit{tasks} during the exploration phase, but for evaluation, the experimenters provide goal images corresponding to tasks like pushing, and stacking that can be performed in the scene. The overall aim is to evaluate goal-reaching behaviors that can be learned by training a goal-conditioned policy described in section~\ref{sec:robotlearning}.\\

\noindent\textbf{Setup.} Our setup is shown in Fig.~\ref{fig:bc}, where we use a Franka Emika Panda robot arm, with an overhead Intel Realsense camera for observations. The robot is controlled through end-effector (EE) control, and the action-space is four dimensional - (x,y,z) position of the EE and opening/closing of the gripper. We place certain objects in the scene, and let the robot interact with them. We choose diverse everyday objects like teddy bears, cloth, ketchup bottle, blocks etc. such that different behaviors like pushing, pick and place, stacking are plausible to emerge through interaction. We reset the scene after $T$ steps of interaction while introducing new objects and/or changing the position of existing objects.  As is standard in evaluating goal-conditioned policies, after training, we measure \% success in reaching a set of test goal-images. For comparison, in addition to goal-sampling from CVAE and Pix2Pix baselines, we consider a curiosity-based exploration baseline~\cite{pathak2017curiosity} for data collection, and follow the same training protocol as our method to train a goal-conditioned policy with the exploration data.\\

\noindent\textbf{Results.} We visualize the progression of exploration during training in Fig.~\ref{fig:explorationviz}, for different episodes corresponding to different sampled goals. Towards the end, we see the evolution of interesting behaviors like grasping and the policy leading to behaviors that reach the goal image. This illustrates that the proposed affordance-guided exploration described in section~\ref{sec:robotlearning} is helpful in training a policy with emergent manipulation capabilities.

After training the policy through affordance-driven exploration, for evaluation we consider a set of goal images, and evaluate the fraction success of the policy in reaching goal configurations. We emphasize that only a \textit{single} goal-conditioned policy is trained and evaluated, but for ease of analysis we show results in Table.~\ref{tb:robot} split across 3 different type of tasks, with the same trained policy. In Table.~\ref{tb:robot}, we show comparisons of the success rates of the two approaches, across three types of tasks - pushing, pick and place, and stacking. For each task, we consider two goal-images, and tabulate \% success across 10 trials. 

In Fig.~\ref{fig:robotexecutions} we visualize some evaluation runs for our method and the curiosity baseline. Please refer to the supplementary video for additional visualizations. We can see an average of 25\% higher success rates compared to the baselines, demonstrating the efficacy of the affordance model for goal-directed exploration.  We observe significantly lower success rates for the CVAE and Pix2Pix baselines, which we believe is because the images generated either do not correspond to realistic reachable goals for the robot or do not change the configuration at all (see examples in Fig. 4), and thus are unable to help in guiding exploration for policy learning. This confirms the generalization of the proposed affordance model trained on diverse passive data, for aiding in goal-directed exploration.


\section{Discussion} 
\label{sec:conclusion}

In this paper we developed a framework for learning visual affordances from passive data such that the learned model can be directly used for goal-directed exploration in unseen scenes. Given an image of an initial scene, our affordance model could generate diverse images corresponding to plausible interactions. Further, its ability to generalize allowed us to directly leverage this affordance model to drive exploration in robot learning. We believe our approach is indicative of the broader potential of large-scale and diverse visual data in developing intelligent agents that can act in generic environments. However, ours is but a first step towards this and we believe several exciting questions remain to be addressed. First, our notion of `visual affordances', while capturing high-level changes, did not model the actions required. In extending this approach to capture `visuomotor affordances', one must overcome the challenges of varying morphology variation across humans and hands, while also developing action representations beyond low-level trajectories. It would also be interesting to generate intermediate `checkpoints' to continuously depict the transition, but these may not be directly useful for robots due to the presence of human hands. Broadly, we are hopeful that extending on our work, there will be future approaches that can leverage increasingly diverse and multi-modal passive datasets for generalization in robot learning, alleviating the necessity of in-domain data collection that has greatly bottle-necked the field. 


\clearpage
\newpage
%
%
\bibliographystyle{splncs04}
\bibliography{references}

 \appendix

\subsection*{Implementation details for the Affordance model and baselines}

\noindent\textbf{VQ-VAE Encoder and Decoder.} We have input image sizes $64\times 64\times 3$ and latent code dimensions $32\times 32\times 1$. The codebook size is 1024, and the codebook dimension is 256. The encoder and decoder architectures are same as~\cite{vqvae}. We use a learning rate of $5e-4$ with ADAM optimizer~\cite{adam}, a batch size of 32, and train for 300K training steps. We use exponential moving average (EMA) to update dictionary items, instead of the commitment loss, with $\gamma=0.99$ as described in Appendix A of~\cite{vqvae}. 

\noindent\textbf{Transformer.}  We follow the ImageGPT~\cite{chen2020generative,lvt} architecture for the auto-regressive Transformer model that predicts goal latent code, given the initial latent code.  The latent code dimensions input are $32\times 32\times 1$ obtained from the VQ-VAE encoder. There are 4 attention heads and 16 attention layers, the vocabulary size is 1024, the embedding size is 512, and the feedforward hidden size is 2048. We use a learning rate of $2e-4$ with ADAM optimizer~\cite{adam}, a batch size of 32, and train for 300K training steps. 

\subsection*{Experiment details for robot learning}

The robot workspace dimensions shown in Fig. 6 of the main paper are approximately $1\text{m}\times 1\text{m}\times 0.4\text{m}$. The observations are obtained from a overhead IntelRealsense camera, and we re-size the images to be of dimensions $64\times 64\times 3$. The policy network takes as input the image observation at time-step $t$, $o_t$ and encodes it with the trained VQ-VAE encoder to a latent $z_t$ of dimension $32\times 32\times 1$. Similarly, the goal image $o_g$ is encoded to $z_g$. The two embeddings $z_t$ and $z_g$ are concatenated, and then there are fully connected layers of dimensions $2048\times 256$, $256\times 64$, $64 \times 4$, with ReLU non-linearities. The action-space is 4 dimensional, and includes the different $x,y,z$ position of the end-effector (i.e. how much to move from the current position) and the final coordinate is for open/close  of the gripper.

\subsection*{ More Qualitative results}

\begin{figure*}[t]
    \centering
    \includegraphics[width=\textwidth]{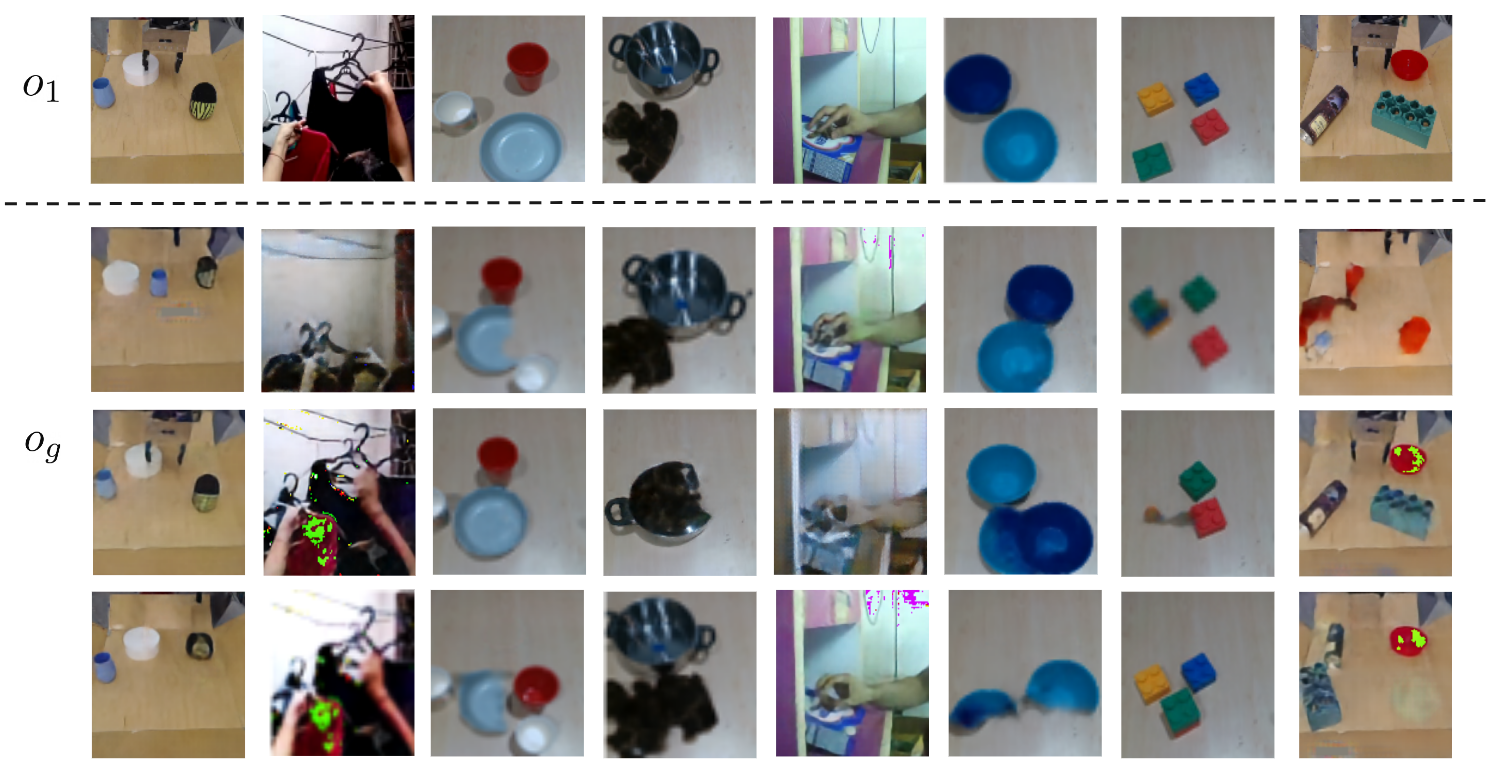}
    \caption{\footnotesize Qualitative analysis of the affordance model, for random initial images. We choose $o_1$ to be a random unseen image, and show randomly sampled affordances $o_g$. We see that the sampled affordances $o_g$ from our model are diverse and correspond to plausible interactions in the scene.}
    \label{fig:qual}
\end{figure*}

In this section, we show more generated affordances for our model for random initial images, in Fig.~\ref{fig:qual}. Visualization of robot exploration as training progresses is in the project website \url{https://sites.google.com/andrew.cmu.edu/affordance-robotics/home} along with an explainer video of motivation, method, and results.

\end{document}

%% file: commands.tex
\usepackage[utf8]{inputenc}
\usepackage[T1]{fontenc}
\usepackage[english]{babel}
\usepackage[]{hyperref}
\usepackage{url}
\usepackage{amsfonts}
\usepackage{nicefrac}
\usepackage{microtype}
\usepackage{amsmath}
\usepackage{amssymb}
\usepackage{mathtools}
\usepackage{subcaption}
\usepackage{ragged2e}
\usepackage{booktabs}
\usepackage{ragged2e}
\usepackage{tikz}
\usepackage{etoolbox}
\usepackage{xspace}
\usepackage{xpatch}
\usepackage{enumerate}
\usepackage{xstring}
\usepackage{setspace}
\usepackage{tabularx}
\usepackage{listings}
\usepackage[capitalise,noabbrev,nameinlink]{cleveref}
\usepackage[useregional=numeric]{datetime2}
\usepackage[ruled,noend]{algorithm2e}
\usepackage{color}
\usepackage{xspace}
\usepackage{pgfplots, pgfplotstable}
\usepackage{xcolor,colortbl}
\usepackage{multicol}
\usepackage{lineno}
\usepackage{wrapfig}

\usetikzlibrary{positioning,arrows.meta,fit,calc,backgrounds}

\definecolor{mydarkblue}{rgb}{0,0.08,0.45}
\hypersetup{%
colorlinks=true,
linkcolor=mydarkblue,
citecolor=mydarkblue,
filecolor=mydarkblue,
urlcolor=mydarkblue}

\creflabelformat{equation}{#2\textup{#1}#3}
\crefname{algocf}{Algorithm}{Algorithms}
\Crefname{algocf}{Algorithm}{Algorithms}

\graphicspath{{figures/}}

\usepackage[skip=1ex]{caption}
\DeclareMathSymbol{\shortminus}{\mathbin}{AMSa}{"39}

\newlength{\philength}
\DeclarePairedDelimiterX{\RoundBrackets}[1]{(}{)}{#1}

\NewDocumentCommand{\pr}{ O{p} r() }{
  \def\prArg{#2}\patchcmd{\prArg}{|}{\mid}{}{}#1\RoundBrackets{\prArg}}
\NewDocumentCommand{\p}{ r() }{\pr[p](#1)}
\NewDocumentCommand{\q}{ r() }{\pr[q](#1)}
\NewDocumentCommand{\Normal}{ r() }{\pr[\operatorname{Normal}](#1)}
\NewDocumentCommand{\Cat}{ r() }{\pr[\operatorname{Cat}](#1)}
\NewDocumentCommand{\Bin}{ r() }{\pr[\operatorname{Bin}](#1)}
\NewDocumentCommand{\Beta}{ r() }{\pr[\operatorname{Beta}](#1)}
\NewDocumentCommand{\Bernoulli}{ r() }{\pr[\operatorname{Bernoulli}](#1)}
\NewDocumentCommand{\Dir}{ r() }{\pr[\operatorname{Dir}](#1)}

\newlength\widthE


%% file: math_commands.tex
\usepackage{amsmath,amsfonts,bm}

\def\eqref#1{equation~\ref{#1}}

\def\1{\bm{1}}

\newcommand{\test}{\mathcal{D_{\mathrm{test}}}}

\DeclareMathAlphabet{\mathsfit}{\encodingdefault}{\sfdefault}{m}{sl}
\SetMathAlphabet{\mathsfit}{bold}{\encodingdefault}{\sfdefault}{bx}{n}

\def\gD{{\mathcal{D}}}

\def\gG{{\mathcal{G}}}

\def\gL{{\mathcal{L}}}

\def\gR{{\mathcal{R}}}

\newcommand{\E}{\mathbb{E}}

